\documentclass[preprint, 10 pt]{elsarticle}
\usepackage{natbib}
\biboptions{authoryear}

\usepackage{graphicx}
\usepackage{hyperref}
\usepackage[utf8]{inputenc}
\usepackage{multirow}
\usepackage{array,multirow}
\usepackage{hyperref}
\usepackage{caption}
\usepackage{subcaption}
\usepackage[legalpaper, margin=1in]{geometry}
\usepackage[table]{xcolor}

\usepackage{amssymb}
\usepackage{amsmath}
\usepackage{xcolor}
\journal{arXiv}
\begin{document}

\begin{frontmatter}


\title{MultiNet with Transformers: A Model for Cancer Diagnosis Using Images}

 \author[address1]{Hosein Barzekar\corref{corresponding author}}
\cortext[corresponding author]{Corresponding author}
\ead{barzekar@uwm.edu}

\author[address1]{Yash Patel\fnref{fn1}}
\fntext[fn1]{The author contributed equally.}
\ead{yspatel@uwm.edu}

\author[address2]{Ling Tong}
\ead{ltong@uwm.edu}

\author[address1,address3]{Zeyun Yu}
\ead{yuz@uwm.edu}

\address[address1]{Big Data Analytics and Visualization Laboratory, Department of Computer Science, University of Wisconsin-Milwaukee, Milwaukee, WI 53211, USA}

\address[address2]{Department of Health Informatics and Administration, University of Wisconsin-Milwaukee, Milwaukee, WI 53211, USA}

\address[address3]{Department of Biomedical Engineering, University of Wisconsin-Milwaukee, Milwaukee, WI 53211, USA}

\begin{abstract}
Cancer is a leading cause of death in many countries. An early diagnosis of cancer based on biomedical imaging ensures effective treatment and a better prognosis. However, biomedical imaging presents challenges to both clinical institutions and researchers. Physiological anomalies are often characterized by slight abnormalities in individual cells or tissues, making them difficult to detect visually. Traditionally, anomalies are diagnosed by radiologists and pathologists with extensive training. This procedure, however, demands the participation of professionals and incurs a substantial cost. The cost makes large-scale biological image classification impractical.  In this study, we provide unique deep neural network designs for multiclass classification of medical images, in particular cancer images. We incorporated transformers into a multiclass framework to take advantage of data-gathering capability and perform more accurate classifications. We evaluated models on publicly accessible datasets using various measures to ensure the reliability of the models. Extensive assessment metrics suggest this method can be used for a multitude of classification tasks.
\end{abstract}


\end{frontmatter}

\section{Introduction}
Breast cancer affects one out of every eight women in their lifetime\citep{tao2015breast}. Preventing breast cancer requires early detection and diagnosis\citep{shaikh2021artificial}. There is currently no magic bullet for the automated detection of breast cancer. For diagnostic and therapeutic purposes, artificial intelligence (AI) has been widely used in biomedical imaging\citep{shin2016deep}. To complete a diagnosis, well-trained specialists usually identify subtleties in cell abnormalities \citep{panayides2020ai}. The capability to identify cell abnormalities requires long-term training. Biomedical imaging has not been widely adopted in some medical facilities, particularly those with limited healthcare resources, due to the high cost of training.\citep{panayides2020ai}. As a cost-cutting measure, AI-based diagnosis has been proposed to reduce costs and increase accuracy. AI approaches have proven to be effective diagnostic tools for biomedical images \citep{yu2018artificial} . Many studies have achieved expert-level disease detection accuracy \citep{rajpurkar2018deep,  tschandl2019expert, haggenmuller2021skin}, including breast cancer\citep{alakhras2015effect, rodriguez2019stand}, skin cancer\citep{brinker2019deep, esteva2017dermatologist}, pneumonia\citep{rajpurkar2017chexnet, basu2020deep}, and hip fracture\citep{gale2017detecting}, as well as other medical fields.

% wang2019pathology, de2020recognition, esteva2017dermatologist, rajpurkar2017chexnet, brinker2019deep, 

Deep learning is an artificial intelligence implementation that can convert large amounts of images into deep domain knowledge for image-based diagnosis \citep{yu2018artificial}. Deep learning incorporates image data into multiple neural network layers to mimic how medical imaging experts learn from experience. To classify disease based on images, these layers convert patterns to digital signals. A specialized diagnostic model labels biomedical images with disease types. Because of the ability to classify biomedical images, deep learning-based classifiers have been widely used in diagnosing diseases from biomedical images\citep{panayides2020ai, mcbee2018deep}. 

Despite deep learning's effectiveness in simulating human learning processes, a number of issues remain unresolved.  First, existing deep learning models are heavily biased toward a specific type of disease, which does not reflect real-world applications. Numerous studies have demonstrated that deep learning can classify numerous types of diseases using binary classification. There is a disconnection between models' binary categorization and an expert's real-world activity of identifying all possible concerns for many diseases based on a single image. This issue may be resolved by the use of multiple classifications for images \citep{miotto2018deep, serag2019translational}.

A deep learning model can perform multiple classifications by acquiring knowledge from images depicting various diseases. However, due to the difficulty of detecting regions of interest in biomedical images, learning to classify multiple diseases needs a deliberate model design. Convolutional neural networks (CNNs) and their modifications have been developed, including VGG-16, VGG-19, Resnet, and Inception. While these models worked admirably for multiple categorizations of general objects, they are unable to meet the demand for multiple classifications of diagnoses inherent in nature. Due to the tiny variance, high dimensionality, and various modalities present in medical imaging, the accuracy of the model of multiple diagnosis disease classification decreases rapidly \citep{simonyan2014very, he2016deep, szegedy2016rethinking}. Current neural network structures are unable to detect small regions of interest from biomedical images. As a result, developing a new neural network structure while keeping unimportant features may be the key to multiple classification of biomedical images.

% We propose that a novel neural network structure which is capable of processing varying image quality, structure, and region-of-interest position can overcome the current limitation. A new well-designed neural network structure has the potential to classify multiple diseases on various types of biomedical images. In this study, we developed two models for classifying multiclass cancers. The first model utilizes the conventional deep learning approach using CNNs and the second model integrates transformers for medical image processing.

Previous studies \citep{che2021constrained, wu2021vision} have shown that a transformer network can successfully address the following biomedical image classification challenges: First, existing AI approaches are not suitable for classifying a variety of disorders. Second, it is indeed not possible to use the current models' architecture to repeatedly classify images of different diseases during training. Our suggested approach not only obtains minor image features that other deep learning models often neglect but also processes images of varying quality, structure, and region of interest.

The purpose of this study is to develop a method for identifying and describing the various anomalies associated with breast cancer. Furthermore, our model improved the ability to detect various types of carcinoma, resulting in a more clinically applicable scenario. Our mode can bridge the gap between Transformer's development and its limited application in medical imaging, specifically in classifying breast cancer sub-types. We used the BreakHis dataset in this study to solve a multiclass classification assignment for eight different types of cancer.

The following contributions are listed:

\begin{itemize}
    \item Initially, we integrated the descriptive capabilities of global and local information of ViT's and CNN's into a single model called MultiNet-ViT. This capability ensures a more distinct feature representation for distinguishing biological image types.
    \item Second, the model incorporates the concept of multiscale image analysis to characterize the details of histopathology images under a microscope, thereby increasing the model's generalizability to different magnification factors.
    
     \item A comprehensive implementation of numerous models is presented. In addition, the model's superior performance and generalizability are compared against additional models.
    
\end{itemize}

\section{Transformer-Based Models}

\subsection{Background}

Transformer\citep{vaswani2017attention}, an alternative to convolutional neural networks,  has dominated the field of natural language processing (NLP), including speech recognition \citep{dong2018speech}, synthesis \citep{li2019neural}, text to speech translation \citep{vila2018end}, and natural language generation \citep{topal2021exploring}. As a example of deep learning architectures, Transformer was first introduced to handle sequential inference tasks in NLP. While recurrent neural networks (RNNs) \citep{graves2013speech} (e.g., long short-term  memory network (LSTM) \citep{sak2014long}) explicitly use a sequence of inference processes, Transformers capture long-term dependencies of sequential data with stacked self-attention layers. In this manner, Transformer is efficient because they solve the sequential learning problem in one shot and effective by stacking very deep models. Several Transformer architectures trained on large-scale architectures have become widely popular in solving NLP tasks such as BERT \citep{devlin2018bert} and GPT \citep{radford2018improving,brown2020language} etc.

Inspired by the success of Transformers in NLP, \citep{dosovitskiy2020image} proposed the Vision Transformer (ViT) by formulating image classification as a sequence prediction task of the image patch (region) sequence, thereby capturing long-term dependencies within the input image. ViT and its derived instances have achieved state-of-the-art performance on several benchmark datasets. Transformers have become very popular across a wide spectrum of computer vision tasks, including image classification \citep{dosovitskiy2020image}, detection \citep{carion2020end}, segmentation \citep{zheng2021rethinking}, generation \citep{parmar2018image}, and captioning \citep{li2019entangled}. Furthermore, Transformers also play an important role in video-based applications \citep{zhou2018end}. Since 2017, Transformers have been used for a variety of computer vision tasks, including general image recognition \citep{touvron2021training,matsoukas2021time}, object detection \citep{carion2020end,zhu2020deformable}, segmentation \citep{ye2019cross}, image classification \citep{che2021constrained},  image super-resolution \citep{yang2020learning}, video interpretation \citep{sun2019videobert,girdhar2019video}, image generation \citep{chen2021pre}, test-to-image integration \citep{ramesh2021zero}, and visual question answering \citep{tan2019lxmert,su2019vl}.

Transformers have recently been adopted in the field of medical image analysis for disease diagnosis \citep{gao2021covid,zhang2021mia} and other clinical purposes. For instance, the works in \citep{costa2021covid,tulder2021multi} utilized Transformers to distinguish COVID-19 from other types of pneumonia using computed tomography (CT) or X-ray images, meeting the urgent need of detecting COVID-19 patients fast and effectively. Besides, Transformers were successfully applied to image segmentation \citep{zhang2021pyramid}, detection \citep{xie2021cotr}, and synthesis \citep{watanabe2021generative}, remarkably achieving state-of-the-art results. Despite the fact that studies have been devoted to customizing Transformers for medical image analysis tasks, such customization has raised new challenges that have yet to be resolved. Many studies have developed efficient Transformers \citep{jaszczur2021sparse,li2022efficientformer,liu2021swin} while maintaining high performances.

Although Transformers performed well in a variety of natural language tasks, their application in biomedical imaging is limited. For example, some preliminary work is shown in the Deformable DETR \citep{zhu2020deformable} model, which is used for object detection. Although many studies are not designed for biomedical images, we believe the attention mechanism also applies to the biomedical imaging field. Max-Deeplab \citep{wang2021max} is the first end-to-end model for panoptic segmentation that demonstrates how Transformers can be used to predict the mask with a category label. A recent co-attention Transformer-based study \citep{chen2021multimodal} demonstrates that several whole slide images can be integrated for patient survival prediction to achieve superior results.Inspired by approaches in Visual Question Answering that can attribute how word embeddings attend to salient objects in an image, the attention mechanism can focus on histology patches that are important predictors of patient survival rate. Other biomedical image classification attempts \citep{chen2021gashis,che2021constrained,zou2021dcet}, segmentation \citep{prangemeier2020attention,yun2021spectr}, and other clinical outcome predictions \citep{chen2021multimodal,zhang2020time,shickel2021multi} have been used. A Transformer network is thought to perform significantly better in multi-model tasks and compensates for the shortcomings of CNN features. The new self-attention mechanism retains long-term features from biomedical image datasets.

\subsection{Vision Transformer(ViT)}

Recently, Vision Transformers (ViT) have achieved highly competitive performance in benchmarks for several computer vision applications, including image classification, object detection, and semantic image segmentation.
Vision Transformer is a model that applies the Transformer to the image classification task and was proposed in 2020 \citep{dosovitskiy2020image}. The model architecture is nearly identical to the original Transformer, but with a twist to allow images to be treated as input in a similar format of natural language processing.

% \begin{figure}
%     \centering
%     \includegraphics[width=\textwidth,height=8cm]{Images/MultiClass/ViT-architecture.png}
%     \caption{ViT Architecture }
%     \label{fig:Vit_architechture}
% \end{figure}

% As shown in \autoref{fig:Vit_architechture},
ViT divides the image into N “patches” of such as 16$\times$16. Since the images are coming in the form of (Height$\times $Width$\times$ number of Channels), $X \in \mathbb{R}^{H \times W \times C}$, they cannot be handled directly by a transformer that deals with (1D) sequences, so it flattens them and makes a linear projection to convert them into 2D patches in the form of $x_p \in \mathbb{R}^{N \times (P^2 .C)}$, Where $N$ is computed using the \autoref{eqNumOfPatches}, and $P$ is the size of each image patch. So each patch can be treated as a token, which can be input to the Transformer.

\begin{equation}
\label{eqNumOfPatches}
\begin{split}
N =\frac{H \times W}{P^2}
\end{split}
\end{equation}

In addition, ViT uses the strategy of pre-training first and then fine-tuning. ViT is pre-trained with JFT-300M, a dataset containing 300 million images, and then fine-tuned for downstream tasks such as ImageNet. ViT is the first pure transformer model to achieve SotA performance on ImageNet, and this has led to a massive surge in research on transformers as applied to computer vision tasks.

Training ViT requires a large amount of data. Transformers are less accurate with less data, but show a quicker accuracy increase with more data, and outperform CNNs when pre-trained on the JFT-300M.

ViT may not benefit from this characteristic because medical datasets are often small. Regarding this problem, a recent investigation by Christos Matsoukas \citep{matsoukas2021time} has compared the advantages of ViT and CNN. They concluded that (1) ViTs pre-trained on ImageNet perform comparably to CNNs when data is limited. (2) Transfer learning favors ViTs when applying standard training protocols and settings. (3) ViTs outperform their CNN counterparts when self-supervised pre-training is followed by supervised fine-tuning. Therefore, these findings suggest that medical image analysis can benefit from the use of ViTs. At the same time, using ViT gained improved explainability because of the attention mechanism, which can highlight the image patches that are important for the classification task. These properties on CNNs are not available. Therefore, we believe ViT is a superior model for biomedical image analysis.

\subsection{Data-efficient image transformers (DeiT)}

One drawback of the ViT transformer is that it does not generalize well when trained on insufficient amounts of data. To overcome the problem that the ViT model must train on large amounts of data. In a paper named “Training data-efficient image transformers $\&$ istillation through attention”, Hugo Touvron, Matthieu Cord, et al. proposed a convolution-free transformer network, DeiT, that achieves top-1 accuracy of 83.1\% on ImageNet with no external data. The training was completed on a single 8-GPU node in less than 3 days. DeiT introduces a new teacher-student strategy specific to transformers that relies on a distillation token, similar to the class token already employed in transformer networks.

Compared to the ViT model which requires hundreds of millions of images, DeiT, however, can be trained easily with approximately 1.2 million images. To attain that goal, they implemented the following strategies:

\begin{itemize}
  \item The first key ingredient of DeiT is its training strategy. Initially, researchers used data augmentation, optimization, and regularization to simulate training on a much larger data set, as done in CNN. 
  \item Further, they modified the Transformer architecture to allow native distillation. (Distillation is a process by which one neural network (the student NN) learns from the output of another network (the teacher NN)). 
  \item They used a CNN as a teacher model for the Transformer.  Using distillation may hamper the performance of neural networks. So, the student model learns from two different sources that may diverge: from a labeled data set (strong supervision) and the teacher.
\end{itemize}

To alleviate the problem, a distillation token is introduced: a learned vector that flows through the network along with the transformed image data and cues the model for its distillation output, which can differ from the token’s class output. This improved distillation method is specific to transformers.

Because of these distillation methods, DeiT performs significantly better than the ViT model on a relatively small biomedical data set. This increases the possibility of using a relatively small dataset.

% One in eight women will develop breast cancer in her lifetime. Preventing the spread of breast cancer requires early detection and diagnosis. As of yet, there is no magic bullet for the automated detection of breast cancer. 
% The goal of this research is to develop a method to identify and describe numerous image anomalies associated with breast cancer. In addition, our model enhanced the ability to detect diverse forms of carcinoma, resulting in a more clinically applicable scenario. Our mode can bridge the gap between Transformer's development and its limited use in medical imaging, particularly in classifying breast cancer sub-types. In this study, we utilized the BreakHis dataset to solve a multiclass classification assignment for eight different types of cancer.

% %\input{02-NMGsStructure.tex}
\section{Methodology}

\subsection{BreakHis dataset}
\autoref{tab:breakHis_Multi} shows the distribution of the BreakHis dataset on multiclass classification. Depending on the microscopic appearance of the tumor cells, both benign and malignant breast cancers can be subdivided into a number of sub-types, each with an unique prognosis and treatment outcome.. The dataset presently comprises four histologically distinct types of benign breast tumors: Adenosis (A), Fibroadenoma (F), Phyllodes Tumor (PT), and Tubular Adenoma (TA); and four histologically unique types of malignant tumors (breast cancer): Ductal carcinoma (DC), Lobular Carcinoma (LC), Mucinous Carcinoma (MC), and Papillary Carcinoma (PC).

\begin{table}[htbp]
  \centering
  \caption{The Distribution of the BreakHis dataset by Magnification Factors and Categories-  Multiclass Classification Setting}
    \begin{tabular}{lllcccc}
    \hline
    \multicolumn{1}{l}{\multirow{2}[0]{*}{Class}} & \multicolumn{1}{l}{\multirow{2}[0]{*}{Sub-Class}} & \multicolumn{4}{c}{Magnifications Factor} & \multirow{2}[0]{*}{Total} \\ \cline{3-6}
          &       & 40X   & 100X  & 200X  & 400X  &  \\  \hline
    \multirow{4}[0]{*}{Benign}  & \multicolumn{1}{c}{F} & 253   & 260   & 264   & 237   & 1014 \\
    & \multicolumn{1}{c}{A} & 114   & 113   & 111   & 106   & 444 \\
          & \multicolumn{1}{c}{TA} & 149   & 150   & 140   & 130   & 569 \\
          & \multicolumn{1}{c}{PT} & 109   & 121   & 108   & 115   & 453 \\ \hline
    \multirow{4}[0]{*}{Malignant} & \multicolumn{1}{c}{DC } & 864   & 903   & 896   & 788   & 3451 \\
          & \multicolumn{1}{c}{LC} & 156   & 170   & 163   & 137   & 626 \\
          & \multicolumn{1}{c}{MC} & 205   & 222   & 196   & 169   & 792 \\
          & \multicolumn{1}{c}{PC} & 145   & 142   & 135   & 138   & 560 \\ \hline
    \multicolumn{2}{l}{Total} & 1995  & 2081  & 2013  & 1820  & 7909 \\ \hline 
    \end{tabular}%
  \label{tab:breakHis_Multi}%
\end{table}%

\subsection{Model Architecture}

\begin{figure}
    \centering
    \includegraphics[width=\textwidth,height=12cm]{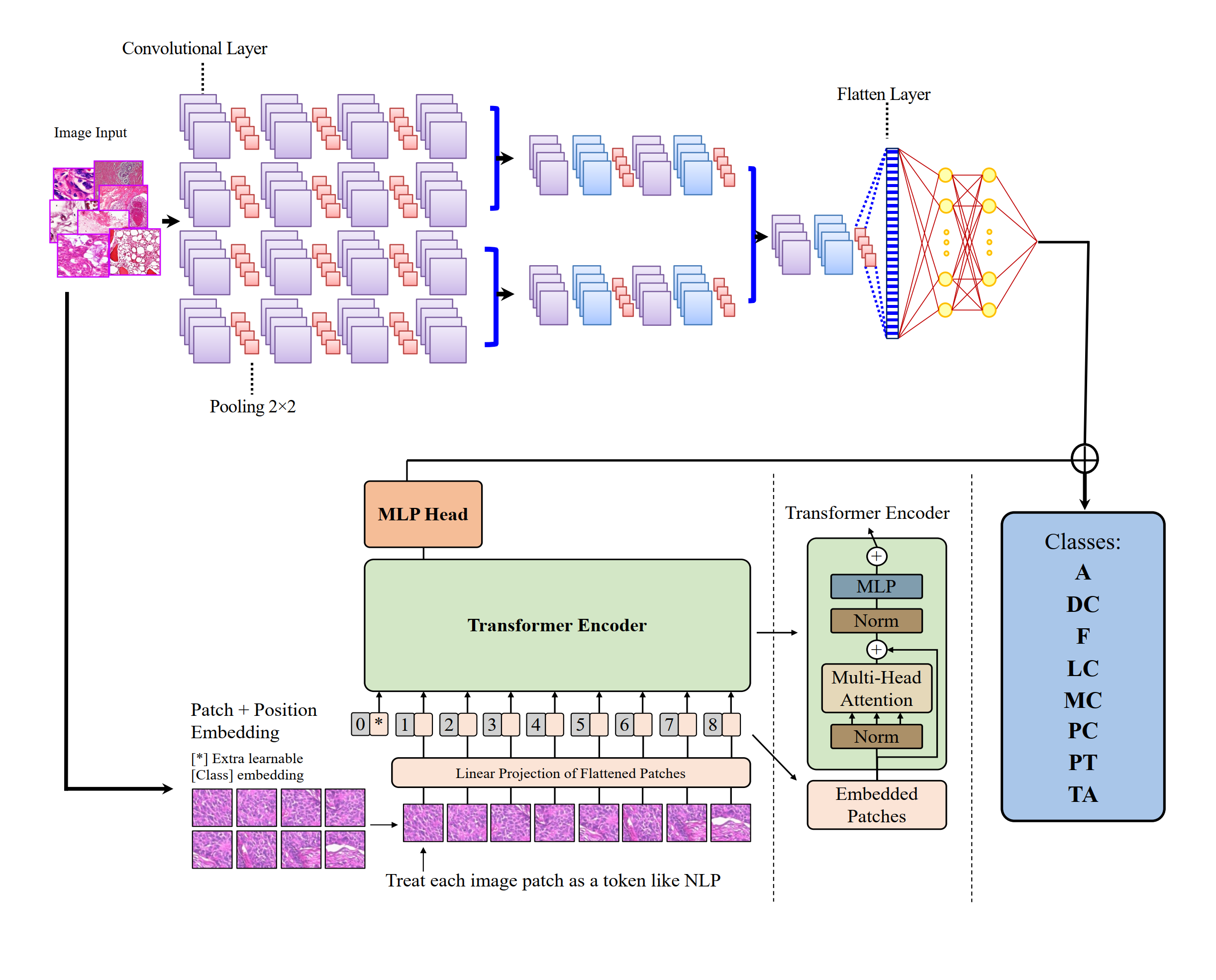}
    \caption{MultiNet-ViT Architecture: Integrating MultiNet with ViT}
    \label{fig:model_vit_Multi}
\end{figure}

The MultiNet model, as shown in \autoref{fig:model_vit_Multi}, entails combining many networks in order to achieve a wide range of aims. Specifically, (1) two transfer learning models, VGG19 and ResNet, are used as the framework's backbone. (2) Since the networks operate in parallel and features are extracted by different networks at different times rather than entering immediately into the fully linked, the inadequacies of one network are offset by another. Then these networks are merged together like the approach used in C-Net \citep{barzekar2022c}. The architecture of the middle, network, and inner network is similar to the C-Net model.

Immediately, Convolutional (Conv) Layers generated by the inner network are fed into a couple of more Conv layers with the following structure; A filter size of 1$\times$1 and stride of 1, the same padding, and 1024 filters followed by an extra 1024 filter with the size of 3$\times$3 and stride of 1, the same padding and a 2$\times$2 max-pooling. These 1024 Conv layers are then convolved through three more Conv layers with 768 as the number of filters for each, however, the filter size for the first Conv layer is 1$\times$1 and the rest are 3$\times$3.

After the preceding Conv layers, the output is fed into a small multi-layer perceptron (MLP) consisting of two linear layers, each followed by a Dropout layer, and a final linear layer with 1024 units and eight outputs (number of classes). All of the linear layers use ReLU activation functions, except for the last one, which uses a Softmax function as shown in \autoref{eqSoftmax}.

\begin{equation}
\label{eqSoftmax}
\begin{split}
\sigma(\Vec{\mathbf{x}})_i =\frac{e^{x_i}}{\sum_{j=1}^C e^{x_j}}
\end{split}
\end{equation}

where $\Vec{\mathbf{x}}$ represents the input vector and C the total number of classes

The MLP head of the MultiNet model is then concatenated, $\oplus$, with the MLP head of the ViT model to classify eight different classes on BreakHis dataset. 

For the proposed model, we use a cross-entropy function as the loss function, as shown \autoref{eqlossFunction} below:

\begin{equation}
\label{eqlossFunction}
\begin{split}
L(\hat{y}, y)=-\left(\sum_{i=1}^{N}y_{i}\log \hat{y}_{i}+(1-y_{i})\log(1-\hat{y}_{i})\right)
\end{split}
\end{equation}

where $y_i$ denotes the $i^{th}$ label $y$ out of N classes, and $\hat{y}$ represents the $i^{th}$ output from the model.

\section{Result and Discussion}

The multiclass classification involving eight classes is conducted. We provide a confusion matrix and following metrics including:  precision and recall, f1 score, and classification accuracy to assess the performance of the model.
Recall is the percentage of images that were successfully classified from the ground truth, whereas precision represents the percentage of images that were accurately classified into that specific predicted class. The F1 score takes into account both precision and recall. The accuracy is the proportion of successfully predicted images (classified images) relative to the total number of predictions.

The model employs a categorical cross-entropy, \autoref{eqlossFunction} loss function.
For loss function minimization, the Adam optimizer is used to fine-tune the weight parameters in order to achieve the best results. $1{e}^{-4}$ is selected as the learning rate, and $\beta_1$, $\beta_2$  is set to 0.9,0.999 respectively. The batches for training, validating, and testing are all set to have a size of 8.

\subsection{ViT-based models}

Leveraging evaluation metrics like accuracy and F1-Score, we present the success of the architecture of ViT-based models in the succeeding results and discussions.

\begin{figure}

        \begin{subfigure}[b]{0.50\textwidth}
        \includegraphics[width=.85\textwidth,height=6.5cm]{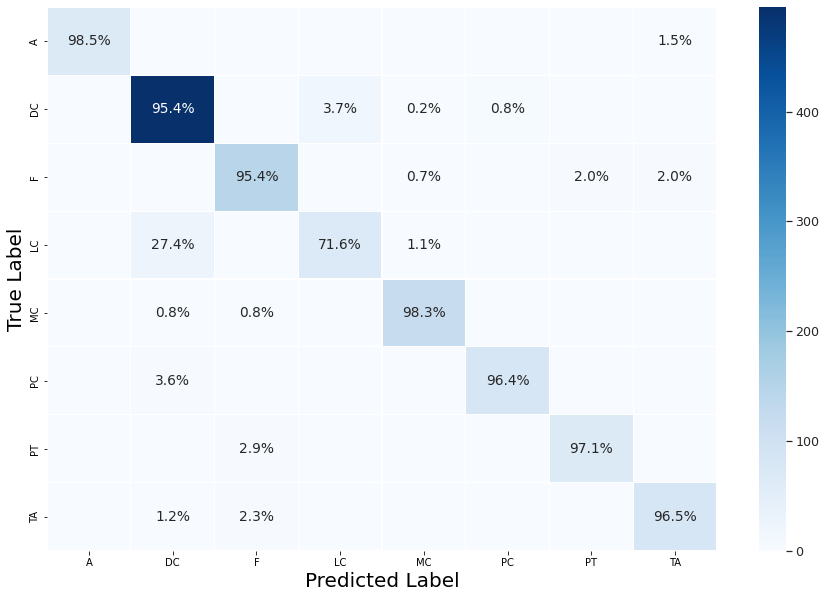}
        \caption{MultiNet}
        \label{fig:cm_vit_multi}
        % \vspace{0.3cm}
        \end{subfigure}%
        \begin{subfigure}[b]{0.50\textwidth}
        \includegraphics[width=.85\textwidth,height=6.5cm]{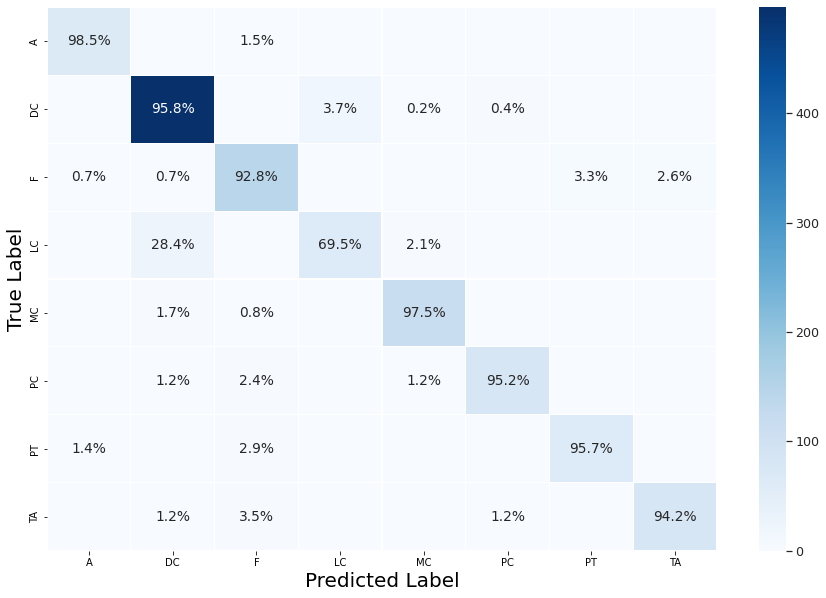}
        \caption{ResNet}
        \label{fig:cm_vit_resnet}
         
        %  \vspace{0.3cm}
        \end{subfigure}% 
       
        \begin{subfigure}[b]{0.50\textwidth}
        \includegraphics[width=.85\textwidth,height=6.5cm]{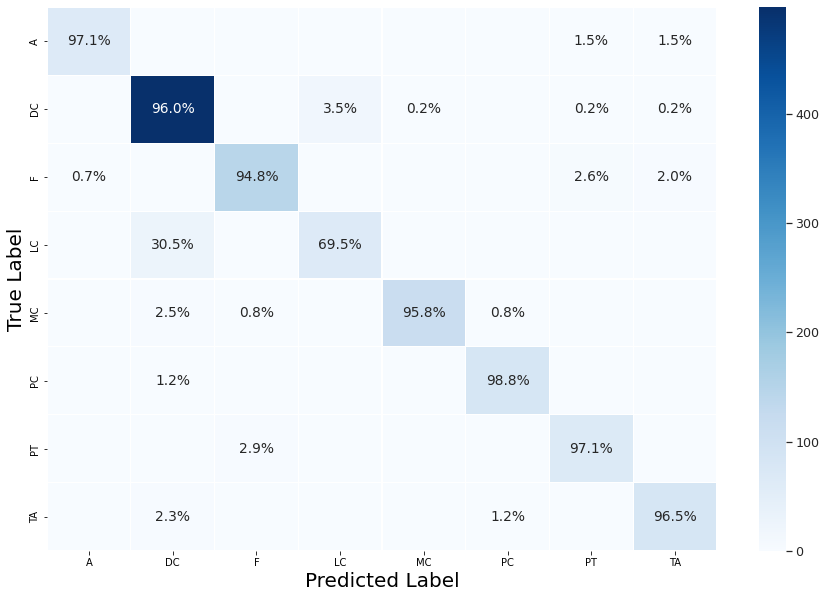}
        \caption{EfficientNet}
        \label{fig:cm_vit_eff}
        % \vspace{0.3cm}
        \end{subfigure}%
        \begin{subfigure}[b]{0.50\textwidth}
        \includegraphics[width=.85\textwidth,height=6.5cm]{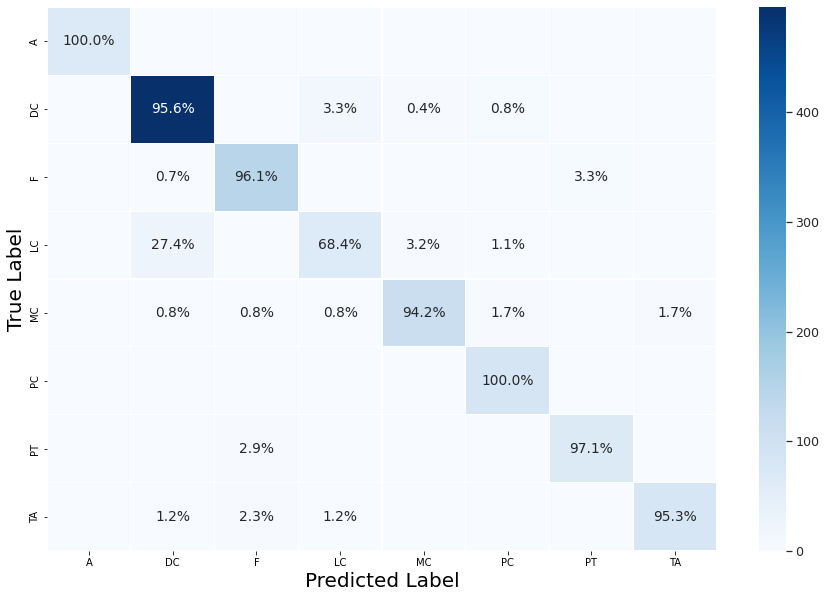}
        \caption{ViT}
        \label{fig:cm_vit}
        % \vspace{0.3cm}
        \end{subfigure}%
        
        \caption{Confusion matrix for the ViT model on BreakHis dataset on the entire dataset including all magnifications 40X, 100X, 200X, 400X}\label{fig:predictions}

\end{figure}

Let us start with ViT model first. The accuracy on each class is shown in \autoref{fig:cm_vit} employing the ViT model alone on the entire dataset combining all the magnification factor including 40X, 100X, 200X, and 400X. As it shows the model achieve 100\% accuracy on adenosis and papillary carcinoma classes.

% VIT model Confusion Matrix
% \begin{figure}
%     \centering
%     \includegraphics[width=.5\textwidth,height=5cm]{Images/MultiClass/cm_Vit_Entire.png}
%     \caption{Confusion matrix for the ViT model on BreakHis dataset on the entire dataset including all magnifications 40X, 100X, 200X, 400X}
%     \label{fig:cm_vit}
% \end{figure}

\autoref{tab:vit_alone} shows the precision, recall, and F1-Score for the ViT model on the BreakHis dataset. ViT achieves 100\% of performance for all the metrics on adenosis class. In addition, the model achieve attain 100\% of recall on papillary carcinoma class. 

% VIT model Precision and Recall
\begin{table}[htbp]
  \centering
  \caption{Precision, Recall, and F1-Score for the ViT model on BreakHis dataset for the multiclass classification} \scalebox{.9}{
    \begin{tabular}{cccc} \hline
    Class & Precision & Recall & F1-Score \\ \hline
    A     &  \textbf{1} &  \textbf{1} &  \textbf{1} \\
    DC    & 0.94  & 0.96  & 0.95 \\
    F     & 0.97  & 0.96  & 0.96 \\
    LC    & 0.77  & 0.68  & 0.73 \\
    MC    & 0.96  & 0.94  & 0.95 \\
    PC    & 0.92  & \textbf{1}     & 0.96 \\
    PT    & 0.93  & 0.97  & 0.95 \\
    TA    & 0.98  & 0.95  & 0.96 \\ \hline
    \end{tabular}%
    }
  \label{tab:vit_alone}%
\end{table}%

\autoref{fig:cm_vit_multi} and \autoref{tab:vit_multi} display the performance of the ViT model integrated with MultiNet model. The model gain 100\% of precision on adenosis class.

% VIT model with MultiNet Confusion Matrix
% \begin{figure}
%     \centering
%     \includegraphics[width=.5\textwidth,height=5cm]{Images/MultiClass/cm_Vit_Multi_Entire.png}
%     \caption{Confusion matrix for the ViT and MultiNet models together on BreakHis dataset on the entire dataset including all magnifications 40X, 100X, 200X, 400X}
%     \label{fig:cm_vit_multi}
% \end{figure}

% VIT model with MultiNet Precision and Recall
\begin{table}[htbp]
  \centering
  \caption{Precision, Recall, and F1-Score for the ViT and MultiNet models on BreakHis dataset for the multiclass classification}  \scalebox{.9}{
    \begin{tabular}{cccc} \hline 
    Class & Precision & Recall & F1-Score \\ \hline 
    A     & \textbf{1}     & 0.99  & 0.99 \\
    DC    & 0.94  & 0.95  & 0.95 \\
    F     & 0.97  & 0.95  & 0.96 \\
    LC    & 0.78  & 0.72  & 0.75 \\
    MC    & 0.98  & 0.98  & 0.98 \\
    PC    & 0.95  & 0.96  & 0.96 \\
    PT    & 0.96  & 0.97  & 0.96 \\
    TA    & 0.95  & 0.97  & 0.96 \\ \hline
    \end{tabular}%
    }
  \label{tab:vit_multi}%
\end{table}%

Precision, recall, and F1-score for the combined ViT with ResNet model on the BreakHis dataset are shown in \autoref{tab:vit_resnet}.

% % VIT model with ResNet Confusion Matrix 
% \begin{figure}
%     \centering
%     \includegraphics[width=.5\textwidth,height=5cm]{Images/MultiClass/cm_Vit_ResNet_Entire.png}
%     \caption{Confusion matrix for the ViT and ResNet on BreakHis dataset on the entire dataset including all magnifications 40X, 100X, 200X, 400X}
%     \label{fig:cm_vit_resnet}
% \end{figure}

Shown in \autoref{fig:cm_vit_resnet} we can observe some extra miss-classification in the ViT with ResNet model compare to ViT with MultiNet model, \autoref{fig:cm_vit_multi}.
% VIT model with ResNet Precision & Recall 

\begin{table}[htbp]
  \centering
  \caption{Precision, Recall, and F1-Score for the ViT and ResNet model on BreakHis dataset for the multiclass classification} \scalebox{.9}{
    \begin{tabular}{cccc} \hline
    Class & Precision & Recall & F1-Score \\ \hline
    A     & 0.97  & 0.99  & 0.98 \\
    DC    & 0.94  & 0.96  & 0.95 \\
    F     & 0.94  & 0.93  & 0.93 \\
    LC    & 0.78  & 0.69  & 0.73 \\
    MC    & 0.97  & 0.97  & 0.97 \\
    PC    & 0.96  & 0.95  & 0.96 \\
    PT    & 0.93  & 0.96  & 0.94 \\
    TA    & 0.95  & 0.94  & 0.95 \\ \hline
    \end{tabular}%
    }
  \label{tab:vit_resnet}%
\end{table}%

The ViT and EfficientNet combined model's confusion matrix is presented in \autoref{fig:cm_vit_eff}. Maximum accuracy of 98\% is attained in the papillary carcinoma class.

% VIT model with EfficientNet Confusion Matrix 
% \begin{figure}
%     \centering
%     \includegraphics[width=.5\textwidth,height=5cm]{Images/MultiClass/cm_Vit_Eff_Entire.png}
%     \caption{Confusion matrix for the ViT and EfficientNet on BreakHis dataset on the entire dataset including all magnifications 40X, 100X, 200X, 400X}
%     \label{fig:cm_vit_eff}
% \end{figure}

In \autoref{tab:vit_eff}, we can see the precision, recall, and F1-score of the combined ViT and EfficientNet model on the BreakHis dataset. The highest value for precision is 99\% and attained on two classes, adenosis and mucinous carcinoma. Furthermore, the model acquires the same performance of recall on papillary carcinoma class. 98\% is the maximum value that the model reaches on F1-Score. 

% VIT model with EfficientNet Precision & Recall 
\begin{table}[htbp]
  \centering
  \caption{Precision, Recall, and F1-Score for the ViT and EfficientNet model on BreakHis dataset for the multiclass classification} \scalebox{.9}{
    \begin{tabular}{cccc} \hline
    Class & Precision & Recall & F1-Score \\ \hline
    A     & 0.99  & 0.97  & 0.98 \\
    DC    & 0.93  & 0.96  & 0.95 \\
    F     & 0.98  & 0.95  & 0.96 \\
    LC    & 0.79  & 0.69  & 0.74 \\
    MC    & 0.99  & 0.96  & 0.97 \\
    PC    & 0.98  & 0.99  & 0.98 \\
    PT    & 0.92  & 0.97  & 0.94 \\
    TA    & 0.94  & 0.97  & 0.95 \\ \hline
    \end{tabular}%
    }
  \label{tab:vit_eff}%
\end{table}%

\subsection{DeiT-based models}
Now, we will examine the performance of the DeiT-based model merged with other models, counting MultiNet, ResNet, and EfficientNet, utilizing assessment criteria such as confusion matrix, precision, recall, and F1-Score.

\begin{figure}
%  \centering
        \begin{subfigure}[b]{0.50\textwidth}
        % \centering
    \includegraphics[width=.85\textwidth,height=6cm]{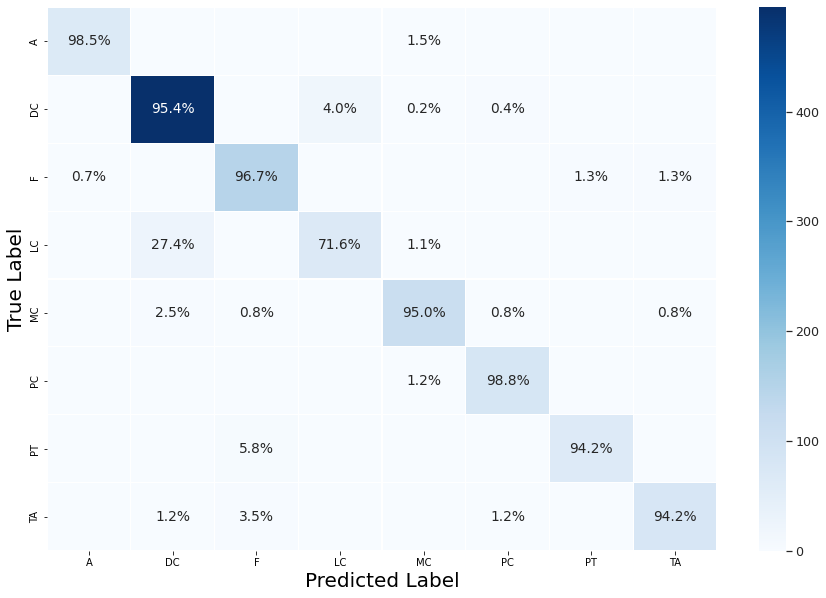}
    \caption{Deit}
    \label{fig:cm_deit}
        % \vspace{0.3cm}
        \end{subfigure}%
        \begin{subfigure}[b]{0.50\textwidth}
    \includegraphics[width=.85\textwidth,height=6cm]{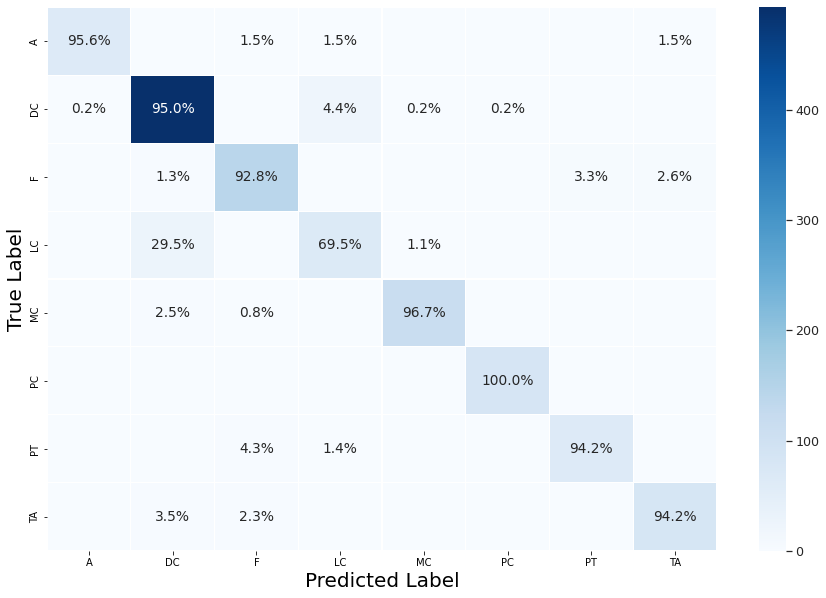}
    \caption{ResNet}
    \label{fig:cm_deit_resnet}
         
        %  \vspace{0.3cm}
        \end{subfigure}% 
       
        \begin{subfigure}[b]{0.50\textwidth}
    \includegraphics[width=.85\textwidth,height=6cm]{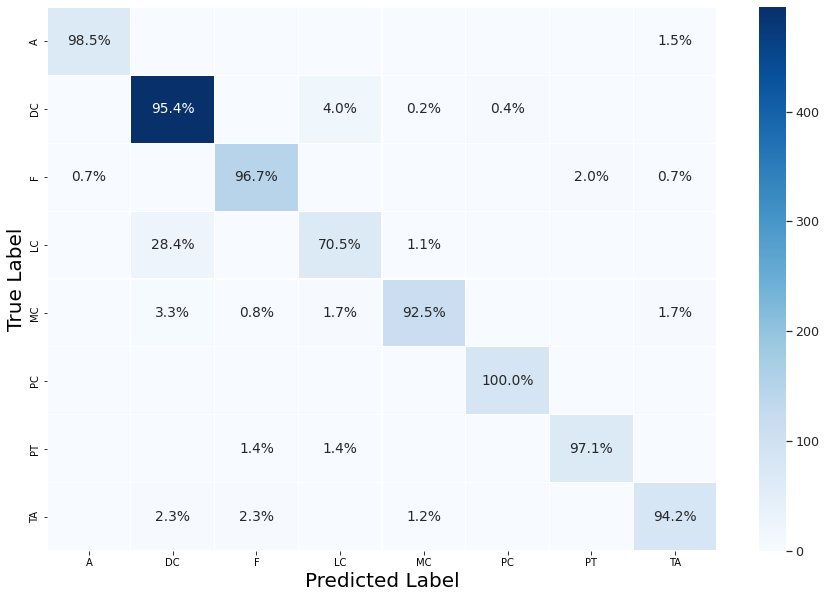}
    \caption{EfficientNet}
    \label{fig:cm_deit_eff}
        % \vspace{0.3cm}
        \end{subfigure}%
        \begin{subfigure}[b]{0.50\textwidth}
    \includegraphics[width=.85\textwidth,height=6cm]{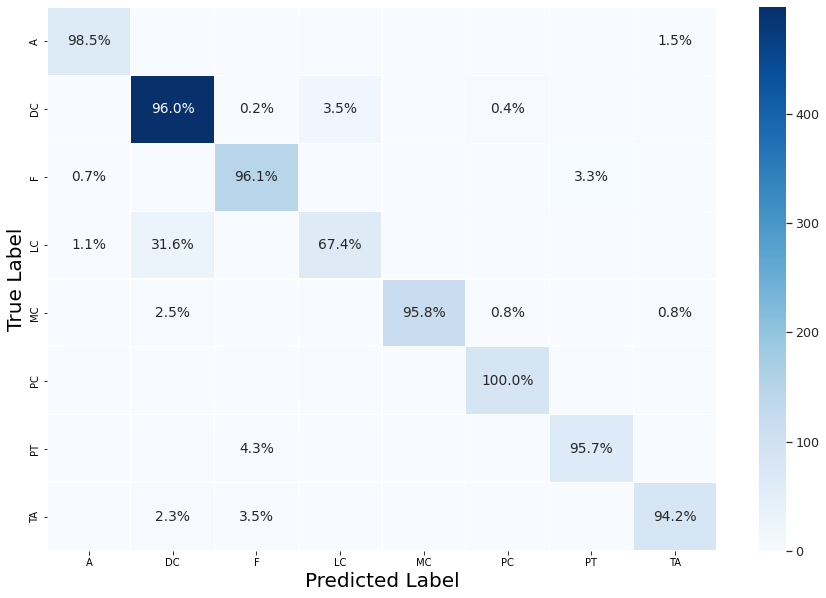}
    \caption{MultiNet}
    \label{fig:cm_deit_multi}
        % \vspace{0.3cm}
        \end{subfigure}%

      \begin{subfigure}[b]{0.50\textwidth}
    %   \centering
    \hspace{4cm}%
    \includegraphics[width=.85\textwidth,height=6cm]{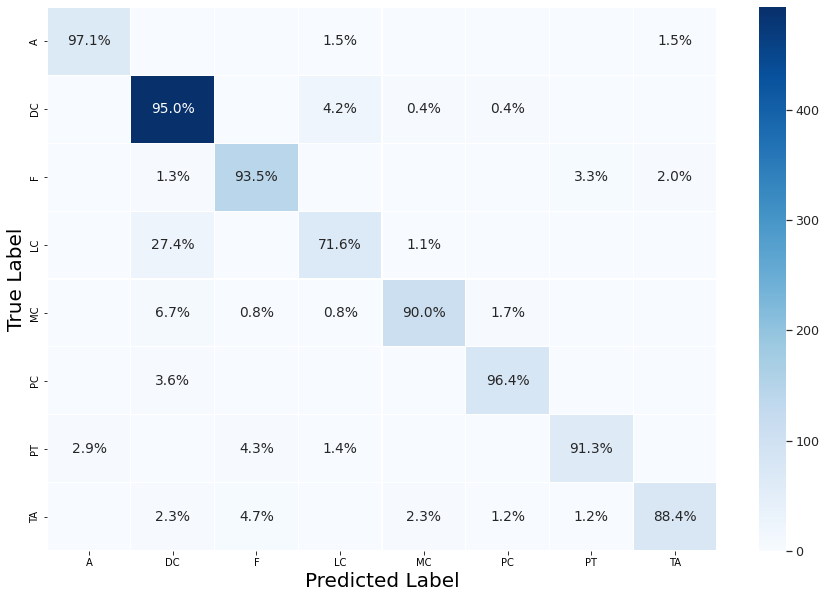}
    
    \centering
    \caption{ViT}
    \label{fig:cm_vit_deit}
        % \vspace{0.3cm}
        \end{subfigure}%
        
    \caption{Confusion matrix for the Deit based models on BreakHis dataset on the entire dataset including all magnifications 40X, 100X, 200X, 400X}\label{fig:conf_Matrices}

\end{figure}

\autoref{fig:cm_deit} illustrates that the papillary carcinoma class has the highest accuracy at 98.8 percent. Except for lobular carcinoma class, all other groups have an accuracy rate of 94\% or above.

% DeiT model Confusion Matrix 

% \begin{figure}
%     \centering
%     \includegraphics[width=.5\textwidth,height=5cm]{Images/MultiClass/cm_Deit_Entire.png}
%     \caption{Confusion matrix for the DeiT on BreakHis dataset on the entire dataset including all magnifications 40X, 100X, 200X, 400X}
%     \label{fig:cm_deit}
% \end{figure}

\autoref{tab:deit_alone} displays the precision, recall, and F1-score achieved by the DeiT model on the BreakHis dataset. Adenosis class yields the highest value, at 99\%, for all the metrics. In addition, similar recall performance is achieved for the papillary carcinoma class.
% DeiT model Precision & Recall 

\begin{table}[htbp]
  \centering
  \caption{Precision, Recall, and F1-Score for the DeiT model on BreakHis dataset for the multiclass classification} \scalebox{.9}{
    \begin{tabular}{cccc} \hline
    Class & Precision & Recall & F1-Score \\ \hline
    A     & 0.99  & 0.99  & 0.99 \\
    DC    & 0.94  & 0.95  & 0.95 \\
    F     & 0.95  & 0.97  & 0.96 \\
    LC    & 0.76  & 0.72  & 0.74 \\
    MC    & 0.97  & 0.95  & 0.96 \\
    PC    & 0.95  & 0.99  & 0.97 \\
    PT    & 0.97  & 0.94  & 0.96 \\
    TA    & 0.96  & 0.94  & 0.95 \\ \hline
    \end{tabular}%
    }
  \label{tab:deit_alone}%
\end{table}%

The confusion matrix for the combined model of DeiT with ResNet is depicted in \autoref{fig:cm_deit_resnet}. In the papillary carcinoma class, the highest possible degree of accuracy is achieved.
% DeiT and ResNet models Confusion Matrix 

% \begin{figure}
%     \centering
%     \includegraphics[width=.5\textwidth,height=5cm]{Images/MultiClass/cm_Deit_ResNet_Entire.png}
%     \caption{Confusion matrix for the DeiT and ResNet model together on BreakHis dataset on the entire dataset including all magnifications 40X, 100X, 200X, 400X}
%     \label{fig:cm_deit_resnet}
% \end{figure}

\autoref{tab:deit_resnet} shows the performance of the DeiT model combined with the ResNet model using Precision, Recall, and F1-Score metrics. The top performance of these metrics all attain by the papillary carcinoma class with following values of 99\%, 100\%, and 99\% for precision, recall and F1-Score respectively. 

% DeiT and ResNet models Precision & Recall 

\begin{table}[htbp]
  \centering
  \caption{Precision, Recall, and F1-Score for the DeiT and the ResNet models on BreakHis dataset for the multiclass classification} \scalebox{.9}{
    \begin{tabular}{cccc} \hline
    Class & Precision & Recall & F1-Score \\ \hline
    A     & 0.98  & 0.96  & 0.97 \\
    DC    & 0.93  & 0.95  & 0.94 \\
    F     & 0.95  & 0.93  & 0.94 \\
    LC    & 0.73  & 0.69  & 0.71 \\
    MC    & 0.98  & 0.97  & 0.97 \\
    PC    & 0.99  & \textbf{1}     & 0.99 \\
    PT    & 0.93  & 0.94  & 0.94 \\
    TA    & 0.94  & 0.94  & 0.94 \\ \hline
    \end{tabular}%
    }
  \label{tab:deit_resnet}%
\end{table}%

In \autoref{fig:cm_deit_eff}, we see the confusion matrix for the DeiT merged with the EfficientNet model, with accuracy at the highest attainable level in the papillary carcinoma class.

% DeiT and EfficientNet models Confusion Matrix 
% \begin{figure}
%     \centering
%     \includegraphics[width=.5\textwidth,height=5cm]{Images/MultiClass/cm_Deit_Eff_Entire.png}
%     \caption{Confusion matrix for the DeiT and EfficientNet model together on BreakHis dataset on the entire dataset including all magnifications 40X, 100X, 200X, 400X}
%     \label{fig:cm_deit_eff}
% \end{figure}

\autoref{tab:deit_eff} illustrates the overall performance of the DeiT with EfficientNet models with respect to the precision, recall, and F1-Score criteria. The adenosis class achieved the highest results across the board (99\%) for each of these indicators of success. Papillary carcinoma class, on the other hand, is slightly easier for the model to identify (recall rate of 100\%) than adenosis class(recall rate of 99\%) and rest of the categories.

% DeiT and EfficientNet models Precision & Recall 

\begin{table}[htbp]
  \centering
  \caption{Precision, Recall, and F1-Score for the DeiT and the EfficientNet models on BreakHis dataset for the multiclass classification} \scalebox{.9}{
    \begin{tabular}{cccc} \hline
    Class & Precision & Recall & F1-Score \\ \hline
    A     & 0.99  & 0.99  & 0.99 \\
    DC    & 0.94  & 0.95  & 0.95 \\
    F     & 0.97  & 0.97  & 0.97 \\
    LC    & 0.74  & 0.71  & 0.72 \\
    MC    & 0.97  & 0.93  & 0.95 \\
    PC    & 0.98  & \textbf{1}   & 0.99 \\
    PT    & 0.96  & 0.97  & 0.96 \\
    TA    & 0.95  & 0.94  & 0.95 \\ \hline
    \end{tabular}%
    }
  \label{tab:deit_eff}%
\end{table}%

\autoref{fig:cm_deit_multi} presents the confusion matrix for the DeiT $\oplus$ MultiNet model. The maximum accuracy is observed in the class papillary carcinomas. In 6 out of 8 categories, the model has a success rate of 95\% or above.

% DeiT and MultiNet Confusion Matrix 

% \begin{figure}
%     \centering
%     \includegraphics[width=.5\textwidth,height=5cm]{Images/MultiClass/cm_Deit_Multi_Entire.png}
%     \caption{Confusion matrix for the DeiT and MultiNet models together on BreakHis dataset on the entire dataset including all magnifications 40X, 100X, 200X, 400X}
%     \label{fig:cm_deit_multi}
% \end{figure}

The DeiT with MultiNet model's overall performance according to precision, recall, and F1-Score is shown in \autoref{tab:deit_multi}. Both the mucinous carcinoma class and the papillary carcinoma class have perfect precision and recall at 100\%. In addition, it has a recall of 99\% for the adenosis class. The F1-Score of the model is 98\% across three different classes (adenosis, mucinous carcinoma, and the papillary carcinoma).

% DeiT and MultiNet models Precision & Recall 
\begin{table}[htbp]
  \centering
  \caption{Precision, Recall, and F1-Score for the DeiT and MultiNet models on BreakHis dataset for the multiclass classification} \scalebox{.9}{
    \begin{tabular}{cccc} \hline 
    Class & Precision & Recall & F1-Score \\ \hline 
    A     & 0.97  & 0.99  & 0.98 \\
    DC    & 0.93  & 0.96  & 0.95 \\
    F     & 0.95  & 0.96  & 0.96 \\
    LC    & 0.78  & 0.67  & 0.72 \\
    MC    & \textbf{1} & 0.96  & 0.98 \\
    PC    & 0.97  & \textbf{1}  & 0.98 \\
    PT    & 0.93  & 0.96  & 0.94 \\
    TA    & 0.98  & 0.94  & 0.96 \\ \hline
    \end{tabular}%
    }
  \label{tab:deit_multi}%
\end{table}%

% DeiT and ViT Confusion Matrix 

\autoref{fig:cm_vit_deit} depicts the confusion matrix for the DeiT paired with the ViT model, with the adenosis class achieving the best accuracy (97\%) of all classes.

% \begin{figure}
%     \centering
%     \includegraphics[width=.5\textwidth,height=5cm]{Images/MultiClass/cm_Vit_Deit_Entire.png}
%     \caption{Confusion matrix for the ViT and DeiT models together on BreakHis dataset on the entire dataset including all magnifications 40X, 100X, 200X, 400X}
%     \label{fig:cm_vit_deit}
% \end{figure}

% DeiT and ViT models Precision & Recall  

We also experimented with an ensemble model for Transformers, which combines the DeiT and ViT models into a unified framework for image classification. \autoref{tab:vit_deit} displays the overall model results with regard to precision, recall, and F1-Score. The adenosis class achieves the highest possible score of 97\% across all metrics.

\begin{table}[htbp]
  \centering
  \caption{Precision, Recall, and F1-Score for the ViT and DeiT model on BreakHis dataset for the multiclass classification} \scalebox{.9}{
    \begin{tabular}{cccc} \hline
    Class & Precision & Recall & F1-Score \\ \hline
    A     & 0.97  & 0.97  & 0.97 \\
    DC    & 0.92  & 0.95  & 0.94 \\
    F     & 0.95  & 0.93  & 0.94 \\
    LC    & 0.73  & 0.72  & 0.72 \\
    MC    & 0.96  & 0.90  & 0.93 \\
    PC    & 0.94  & 0.96  & 0.95 \\
    PT    & 0.91  & 0.91  & 0.91 \\
    TA    & 0.95  & 0.88  & 0.92 \\ \hline
    \end{tabular}%
    }
  \label{tab:vit_deit}%
\end{table}%

\begin{figure}
    \centering
    \includegraphics[width=.9\textwidth,height=8cm]{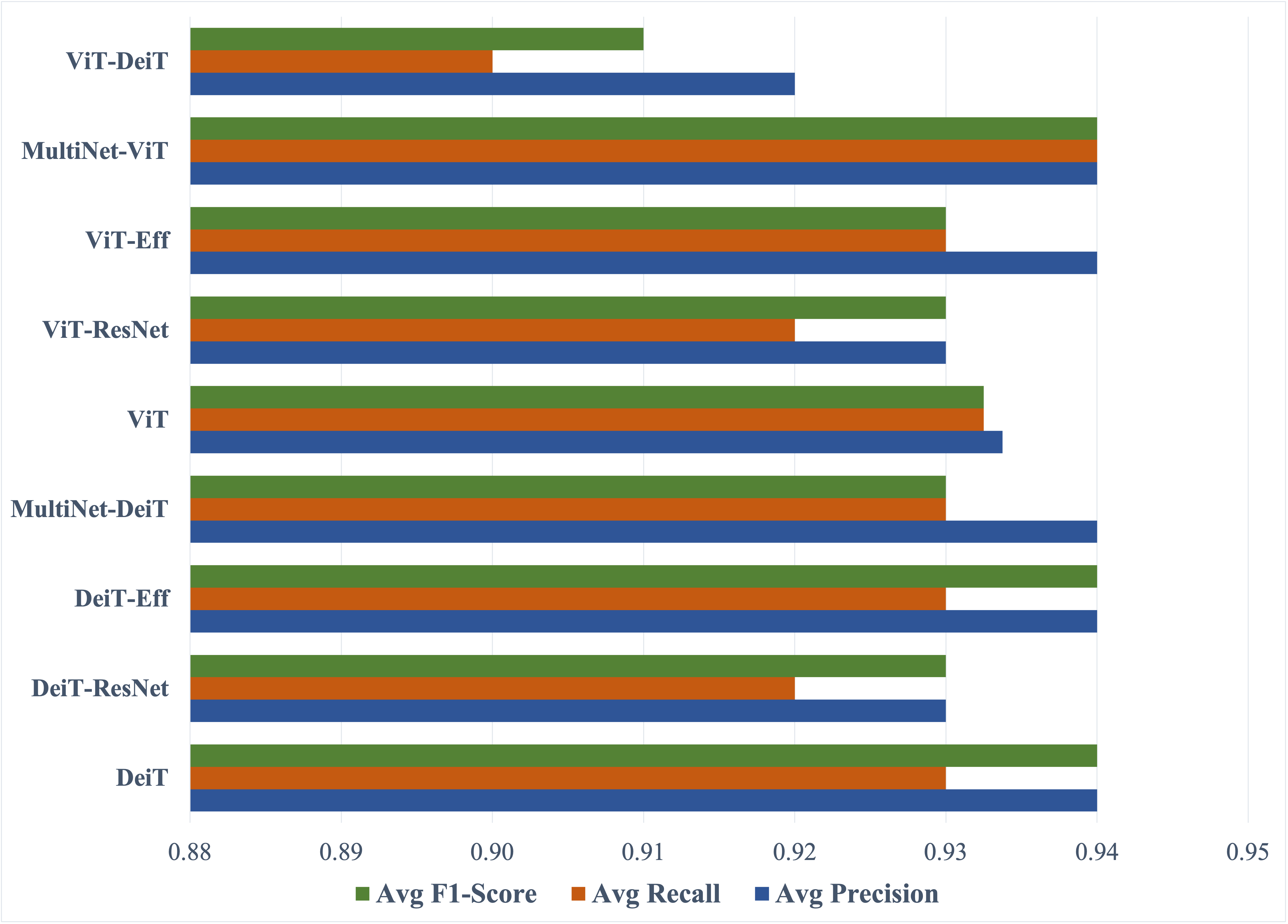}
    \caption{Average of the Precision, Recall and F1-Score of all the models for the multiclass classification on BreakHis dataset. Eff(EfficientNet)}
    \label{fig:avg_prRe_Multiclass}
\end{figure}

\autoref{fig:avg_prRe_Multiclass} displays the mean precision, recall, and F1-Score of all models utilized to conduct the experiments. As observed, MultiNet-ViT gets the best performance among all metrics at 94\%. The proposed model outperforms all other transformer-based combinations. MultiNet-ViT offers a significant degree of generalizability for multiclass classification. Using multiple networks powered by transformers can result in a more reliable and efficient solution.

%Average of Precision and Recall and F1-Score of all the model 

\section{Conclusion}

Breast cancer is the most common type of cancer in females. Early detection and identification can stop the further development of breast cancer. The ideal application in automated breast cancer detection is still a work in progress.
In this research, we propose a couple of novel Neural Network models to diagnose different cancer types. The architectures was tested on BreakHis dataset. To assure reliability, the proposed model has been evaluated using several metrics.

 MultiNet-ViT is suggested to  identify breast cancer on histological images. Taking into consideration both the global and local associations of images in a broad context, this approach combines the strengths of the traditional CNN model for extracting local information and the latest transformer model for collecting long-range correlation. The model was evaluated on the BreakHis dataset of breast cancer histopathology, where it showed an average of 94\% precision, recall, and F1-Score. Results from our trials definitively show the transformer's potential in the multiclass classification tasks of breast cancer sub-types. So as to guarantee the model's generalization abilities, we have also expanded our detection goal to encompass all magnification factors. Finally, in order to gauge the efficacy of our model, we compared it to that of competing models. The proposed models improved the detection capability for different types of carcinomas, resulting in a more applicable scenario for clinical application. Our model is able to bridge the gap between transformer's development and its limited use in medical imaging, particularly in breast cancer multiclass classification.

Transformers without transfer learning demand an excessive amount of data, training, and processing resources. Accordingly, it is recommended to utilize transfer learning, particularly in medical image processing. Future studies may find a similar strategy works well for other computer vision tasks, such as detection, segmentation, localization, etc. In addition, the architecture has not been applied to three-dimensional volumes, therefore any such future efforts would be quite informative.

\bibliographystyle{apalike}
\bibliography{References}

\end{document}